# Prediction of *MET* Overexpression in Non-Small Cell Lung Adenocarcinomas from Hematoxylin and Eosin Images

## Author List & Affiliations

Kshitij Ingale[1], Sun Hae Hong[1], Josh S.K. Bell[1], Abbas Rizvi[1], Amy Welch[1], Lingdao Sha[2], Irvin Ho[1], Kunal Nagpal[1], Aicha BenTaieb[3], Rohan P Joshi[1*], Martin C Stumpe[1]

[1]Tempus Labs Inc, Chicago, IL

[2]Work done at Tempus Labs Inc, current affiliation: Amazon

[3]Work done at Tempus Labs Inc, current affiliation: Genentech

Corresponding Author:

Rohan P Joshi

600 W Chicago Ave Ste #510, Chicago, IL 60654

(833) 514-4187

rohan.joshi@tempus.com

## Competing Interests Statement

All authors were employees and shareholders of Tempus Labs at the time this work was done.

# Abstract

MET protein overexpression is a targetable event in non-small cell lung cancer (NSCLC) and is the subject of active drug development. Challenges in identifying patients for these therapies include lack of access to validated testing, such as standardized immunohistochemistry (IHC) assessment, and consumption of valuable tissue for a single gene/protein assay. Development of pre-screening algorithms using routinely available digitized hematoxylin and eosin (H&E)-stained slides to predict MET overexpression could promote testing for those who will benefit most. While assessment of MET expression using IHC is currently not routinely performed in NSCLC, next-generation sequencing is common and in some cases includes RNA expression panel testing. In this work, we leveraged a large database of matched H&E slides and RNA expression data to train a weakly supervised model to predict *MET* RNA overexpression directly from H&E images. This model was evaluated on an independent holdout test set of 300 over-expressed and 289 normal patients, demonstrating an ROC-AUC of 0.70 (95th percentile interval: 0.66 - 0.74) with stable performance characteristics across different patient clinical variables and robust to synthetic noise on the test set. These results suggest that H&E-based predictive models could be useful to prioritize patients for confirmatory testing of MET protein or *MET* gene expression status.

# Introduction

Non-small cell lung cancer (NSCLC) is one of the most common cancer types in the world, resulting in poor outcomes for many patients [1]. Precision medicine applications aimed at biomarker testing and targeted therapy are a continued area of active research [2].

The *MET* oncogene is a therapeutically actionable target in NSCLC and encodes a receptor tyrosine kinase involved in cell growth and survival. Protein overexpression, gene amplification, and gene mutation have been identified as mechanisms for *MET* gene de-regulation [3]. Gene amplification of *MET* is correlated with shorter survival and has also been identified as a drug-targetable event in NSCLC [4,5]. The antibody-drug conjugate telisotuzumab vedotin has been granted FDA Breakthrough Therapy Designation [6] for patients whose tumors overexpress MET protein as determined by immunohistochemistry (IHC). *MET* DNA copy number has been found to have poor concordance with MET protein expression [7,8]. However, RNA expression could correlate with MET protein expression [9], providing a biomarker that would allow simultaneous probing of other actionable alterations in NSCLC by combined DNA and RNA next-generation sequencing (NGS) [10].

While panel-based DNA NGS is routinely performed for detection of actionable mutations in NSCLC, IHC testing for MET is not currently part of clinical workflows. In contrast, clinical diagnostic workflows for NSCLC typically involve creating slides stained with hematoxylin and eosin (H&E). With the rise in digital pathology applications, slides are increasingly being digitized to create high-resolution scans, termed as whole-slide images (WSIs). Assuming that molecular alterations of interest lead to distinct visual characteristics in the H&E slide morphology, the application of machine learning on WSIs can be used to predict the presence of molecular biomarkers. Such predictive models could be used to prioritize tissue for subsequent molecular testing for targetable biomarkers.

Building these algorithms requires the use of gigapixel WSIs that are paired with a single molecular label. Due to the lack of human understanding of genotype image-phenotype connections for many molecular labels, pixel-level annotations are typically not available for this type of prediction. Weakly supervised methods like multiple instance learning address this issue by allowing training with gigapixel images to predict a single slide label [11,12]. These algorithms typically require large datasets of WSIs with corresponding labels and different data modalities including DNA-based variant calling or RNA-based gene expression, which are often limiting factors for their development. Algorithms also need to be carefully validated with regards to their real-world deployment use-cases. Their performance might be affected by deviation of patient characteristics and scanning devices between real-world data and training datasets. Understanding the robustness of algorithms to real-world scenarios such as changes in slide colors or patient characteristics is critical. Additionally, it is also important to confirm stable performance over time after model development.

In this work, we use a large dataset of NSCLC adenocarcinoma H&E WSIs and corresponding RNA expression labels to develop and validate a model to predict *MET* overexpression status, accounting for color variation, scanner type, and patient subgroups.

## Methods

### Cohort curation

For model development, a dataset of 1,367 H&E slides (Supplementary Table 1) digitized with a Philips UFS scanner was curated from a de-identified database of patient records (Tempus Labs, Chicago, IL). To be included in the cohort, patients were required to have a diagnosis of NSCLC adenocarcinoma. Cases that underwent bone marrow core biopsy, fine needle aspirate, fluid aspirate, skin punch, and/or venipuncture were excluded to ensure relative histological architecture was not damaged during sample extraction. Furthermore, digitized images underwent a manual quality control process to ensure high quality images were selected for modeling. As a part of this, trained pathologists removed cytology specimens and slides with less than 100 viable tumor cells as biomarker status is difficult to determine in such cases [13]. Slides with small out-of-focus areas were included in the cohort to ensure the trained models were robust to this situation, which commonly occurs due to slide preparation factors such as variable tissue thickness. Conversely, grossly out-of-focus images were excluded from the cohort since they are rare and not meaningful for training or inference. Across all manual quality control, <2% of total slides were removed. The model was evaluated on a test set of 589 internally stained H&E slides (Supplementary Table 1) digitized with a Philips Ultra Fast Digital Pathology Slide Scanner (Philips, Amsterdam, NL), a test set of 519 internally stained H&E slides (Supplementary Table 4) digitized with an Aperio GT 450 Digital Pathology Slide Scanner (Leica Biosystems, Wetzlar, Germany), and a test set of 4,331 H&E slides (Supplementary Table 4) stained internally and at external labs digitized using Philips or Aperio scanners.

Generating labels for training

*MET* DNA-amplified cases represent a subset of cases in which *MET* RNA expression is expected to be high. In contrast, *MET* copy number-neutral cases could have high *MET* RNA expression through other mechanisms (for example, epigenetic changes). To determine a suitable threshold to define high RNA expression using RNA-seq data, we compared *MET* RNA expression transcripts per million (TPM) values to *MET* DNA copy number as assessed by our clinical NGS pipeline for thousands of cases. As a sample may have different copy numbers in segments of the *MET* gene, we computed an average of the copy numbers weighted by the segment length. This was used to group patients into amplified (segment-weighted copy number greater than or equal to 5) and wild type (segment-weighted copy number equal to 2). We compared RNA expression values for these two groups and plotted a receiver operating characteristic curve (ROC) (Supplementary Figure A2). We optimized for a high sensitivity threshold given the expected concordance of *MET* DNA-amplified and *MET* RNA expression-high labels. We found that a threshold of 8 log TPM separated amplified and neutral cases with a specificity of 80% and sensitivity of 92%, consistent with our desire for a high sensitivity operating point. The ground truth label for RNA expression was obtained using this threshold and later used for training an attention-based multiple instance learning model (Figure 1b). The cohort for model development was subsampled to enrich for cases belonging to the positive class (Supplementary Figure A1).

## Dataset preprocessing

For each slide, tissue-containing regions were segmented using a U-Net model [14], and 224x224 pixel tiles were generated from the tissue regions at 5x, 10x and 20x magnification. Some slides contained pathologist marker annotations that could lead to the model learning marker presence instead of the target. A U-Net-based model was used to segment marker regions and exclude tiles that overlap with marker regions.

## Model Training and Evaluation

Attention-based multiple instance learning models similar to Ilse et al. [15] were trained using the tiles from digitized H&E images and their corresponding RNA labels. Within this model, a ResNet18 [16] model initialized with weights pre-trained on a histopathological task was used as a feature extraction module. This model was then trained on the images with Adam optimizer [17] using a learning rate of 1e-5 and weight decay of 1e-2. We used cross-entropy loss to train the models. The raw RNA expression values measured in TPM were transformed using a log function followed by a sigmoid function to obtain labels using the following equation. We observed best performance when the sigmoid function was scaled by a factor of 2.

$$Smooth\ label = sigmoid(2 * (log_2(TPM + 1) - 8))$$

We used 5-fold cross-validation to obtain robust model performance estimates for finding the best tile magnification and label strategy (see effect of model development parameter choices under the Results section). Out of 5 folds, 3 folds were used for training, 1 fold was used for selecting training epoch with early stopping, and 1 fold was used for calculating generalization performance. In order to train the final model, the dataset was reshuffled followed by creating training and model selection sets. The model selection set was used to select epoch through early stopping as well as selecting the best learning rate and weight decay (best learning rate was 1e-5 and weight decay was 1e-2).

In order to get an idea of dataset sufficiency, we compared performance as a function of dataset size by using increasingly smaller subsampled fractions of data for training. Repeated random subsampling was performed for subsets to account for variance in sampled cases.

The final retrained model was then evaluated using an independent holdout test set digitized with a Philips UFS scanner and another holdout test set digitized with an Aperio GT 450 scanner. In addition to this, the model was evaluated to another test set of specimens collected and digitized after 6 months from the model development set (Supplementary Table 4). This test set had cases stained internally and at other external labs that were scanned with Philips and Aperio scanners. Scanning devices can have different color profiles identified by ICC profile. We applied ICC profile transformation to translate images to standard (sRGB) color space as it has been shown to improve generalization [18]. The model was also evaluated for robustness to synthetic noise to brightness, contrast, hue, and saturation of input images. The perturbation factor was sampled from increasingly wider intervals (Figure 5b) and repeated multiple times to capture variance in model performance for the range of noise. Model performance was tracked using metrics including ROC-AUC, average precision, sensitivity, specificity, precision, F-1 score and Cohen's Kappa between model predictions and ground truth labels.

### Attention scores analysis

Tile histology classes of tumor, stroma, epithelium, necrosis, immune, and artifact were predicted by a custom tile histology classification pipeline [19]. Histology fractions in all tiles were determined in each slide as a baseline. Then in each slide we calculated histology fractions in the high-attention tiles, tiles accounting for the top 10% of the cumulative attention probability. We employed Wilcoxon signed-rank test with Benjamini/Hochberg False Discovery Rate (FDR) correction to assess i) Tile class fraction difference between high attention tiles and all tiles and ii) Over-representation of tumor tile classes over other tile classes in high-attention tiles.

### Embedding analysis

Tile feature embedding vectors were extracted from the average pool layer of the Resnet18 backbone. Tiles were pooled over all slides in the Philips UFS scanner holdout test set (n=588). A UMAP [20] model was trained on the pooled tile embeddings to reduce 512 dimensional embeddings for visualization in 2D plot. Tiles were grouped to buckets of percentiles of the raw attention scores. Tile classes were assigned by the custom tile histology classification pipeline.

# Results

## Weakly supervised model learns *MET* expression signal

We trained weakly supervised attention-based multiple instance learning models to predict *MET* RNA overexpression. The model achieved a mean ROC-AUC of 0.74 (95th percentile interval: 0.67 - 0.81) in cross-validation (Figure 2a). We observed a significant correlation between RNA expression TPM values and model predictions (Pearson's r = 0.43, $p < 0.001$; Figure 2b).

We examined the role of magnification level for prediction of *MET* RNA overexpression as lower magnifications capture tissue architecture while higher magnifications capture cellular morphologies. We found that models trained with tiles at 20x magnification outperformed the other models ($p$ = 0.02 and $p$ = 0.003 for comparing against 5x and 10x tiles, respectively; Figure 2c), with a mean ROC-AUC of 0.72 (95th percentile interval: 0.71 - 0.74). We also investigated modeling *MET* RNA expression as a continuous measurement ("smoothed labels") or a binary label of overexpressed or underexpressed using a DNA amplification-defined threshold (Supplementary Figure A2, Methods). We found that models trained with binary labels trended towards lower ROC-AUCs (mean = 0.72 vs mean = 0.74, $p$ = 0.11). Additionally, the predicted probabilities from the binary-label model had lower correlation with RNA overexpression values (r = 0.41 vs r = 0.43, Supplementary Figure A3), suggesting poorer model calibration.

Because multiple instance learning with neural networks requires large training datasets, we wanted to determine if addition of cases to our training dataset would yield meaningful increases in performance. We therefore analyzed the relationship of test performance with an increasingly smaller subsample of positive cases available for training. Generalization performance increased as dataset size was increased as expected (Figure 2d). The data titration trend suggested that adding more cases to our dataset would bring marginal improvements in performance.

## Model generalizes to an unseen test set

The final model performance was evaluated on an independent holdout test set that was not used previously during model development (Supplementary Table 1). We set an operating point using a portion of the training set and targeted a sensitivity of 80%. The model had an ROC-AUC of 0.70 (95th percentile interval: 0.66 - 0.74; Figure 3a) and average precision of 0.69 (95th percentile interval: 0.65 - 0.74; Figure 3b) on this test set. We also compared the predicted probabilities from the model with RNA expression values and observed a significant Pearson correlation of 0.4 ($p < 0.001$; Figure 3c). Performance metrics on the holdout set are summarized in Figure 3d.

In addition to identifying meaningful signals from data, machine learning models may learn biases prevalent in data too. Biases can be concerning for real-world deployment of a model and lead to unpredictable performance for particular subgroups. We investigated the relationship between performance and gender, race, smoking status and stage at diagnosis, and tissue-related extraction method or tissue site. We observed a statistically significant difference in ROC-AUC between cases extracted using methods that yield larger sections of tissue, such as surgical resection, and those that yield smaller sections like core needle biopsy (large specimen: ROC-AUC: 0.79, 95th percentile interval: 0.72 - 0.85 vs small specimen: ROC-AUC: 0.67, 95th percentile interval: 0.61 - 0.72; $p = 0.004$). We did not detect meaningful differences in performance in the other clinical subgroups with our limited sample size (Figure 4a).

Because our model is trained with weak supervision, the model might learn features corresponding to a co-occurring mutation status instead of *MET* status. We therefore compared model performance for predicting *BRAF*, *EGFR*, *KRAS*, and *TP53* gene mutations, which are commonly observed in NSCLC adenocarcinoma. Using the output of the *MET* model as a score, we observed that ROC-AUCs for these molecular labels were significantly different from the ROC-AUC for *MET* status (Figure 4b), suggesting that the model has learned features optimized for *MET* status prediction and not for these other mutations.

# Model performs consistently across scanners, staining, and time

One of the major hurdles in machine learning model deployment is the wide variety of data encountered in the real world. Some of this variability arises from differences in lab staining protocols and scanning devices. The variance in real-world data images could manifest as significant visual color differences. We therefore evaluated the sensitivity of model performance to increasing levels of perturbations of brightness, contrast, hue, and saturation (see Methods). We found that ROC-AUC and average precision were similar after perturbations, with slight decreases for very high perturbation (Figure 5a, additional metrics in Supplementary Figure A5a). However, performance of operating point-based metrics like sensitivity and specificity was generally stable to level 5 perturbation, and subsequently diverged as the magnitude of perturbation increased. Together these results suggest that while the rank-ordering of predictions from the model is robust to color variation, significant shifts in model output probabilities occur (Supplemental Figure A5b).

## Performance on a different scanner

In a deployment scenario machine learning models could encounter images scanned with a different scanning device. This could result in a shift in deployment data distribution relative to training data distribution. We observed a slight decrease in mean ROC-AUC in cross-validation for models trained only on Philips UFS-scanned images and evaluated on Aperio GT450-scanned images compared to models trained on Philips UFS-scanned images and evaluated on Philips UFS-scanned images (ROC-AUC 0.70 vs 0.74, *p* = 0.02; Supplementary Figure A5). We also evaluated our model that was trained with only Philips UFS-scanned images on a separate holdout test set of images that was scanned using an Aperio GT450 (Supplementary Table 4). ROC-AUC and AUPRC, two rank-based metrics, were similar between test sets scanned with Philips and Aperio (Figure 6). However, we observed lower sensitivity and higher specificity for the Aperio test set inference compared to the Philips test set. The observed differences between rank-based and operating point-sensitive metrics are consistent with a noticeable shift in the distribution of predicted probabilities between the two scanners (Supplementary Figure A6).

Performance on temporal holdout test set and on external staining patterns

The performance of machine learning models can drift over time due to ongoing changes in characteristics of input data relative to training data used for model development, such as changing staining protocols and scanning device calibration. To evaluate our model's robustness to drift, we curated a test set of patient samples imaged and sequenced starting 6 months after our training and holdout test sets and continuing for a period of 10 months (Supplementary Table 4). This test set included tissue slides stained at our institution as well as other pathology labs and slides scanned with Philips and Aperio scanners. We observed a stable performance on cases scanned with a Philips scanner, with the model yielding an ROC-AUC of 0.71 (95th percentile interval: 0.60 - 0.81; Figure 7a) on cases stained at external labs and an ROC-AUC 0.71 (95th percentile interval: 0.67 - 0.75; Figure 7b) on cases stained internally. We noted a non-significant decrease in performance on cases scanned with an Aperio scanner, with an ROC-AUC of 0.67 (95th percentile interval: 0.61 - 0.74; Figure 7c) on cases stained at external labs and an ROC-AUC 0.68 (95th percentile interval: 0.66 - 0.71; Figure 7d) on cases stained internally. This further confirms that model performance is indistinguishable between scanner types.

# Model gives tumor-containing tiles high importance for predictions

To assess if the model is making predictions based on biologically relevant information, we applied a custom histology-classifying neural network to the same tiles that were input into the *MET* predictive model (Figure 1b). This network classified each tile into one of 6 histology classes: tumor, stroma, epithelium, necrosis, immune, or artifact. For each tile, we also extracted a corresponding attention score from the *MET* predictor model. This attention score provides a measure of tile importance for the model's prediction.

We calculated the fraction of histology classes in all tiles and the fraction of histology classes in tiles given high-attention importance by our model (see Methods, Figure 8a). Tumor class was more enriched in high-attention tiles ($p < 0.001$) while stroma, epithelium, necrosis, and immune classes were reduced (Supplementary Table 2). Within the high-attention tiles for each slide, the fraction of tumor class tiles was higher than all other classes ($p < 0.001$ for artifacts, epithelium, immune, necrosis, and stroma classes, Wilcoxon signed-rank test with Benjamini/Hochberg FDR correction; Supplementary Table 3). These results suggest the model places importance on tumor-containing tiles for prediction, consistent with their biological relevance for *MET* overexpression.

To further understand the mechanism of the model, for each tile we additionally extracted tile embedding vectors (see Methods). Tile embedding vectors represent information the trained model derives from a tile image. High-dimensional tile embedding vectors were projected to two dimensions using UMAP dimensionality reduction to visualize the embedding space. In the embedding space, tiles of the same histology class generally clustered together (Figure 8b), suggesting the model learned to embed information related to histological class. Consistent with tumor class enrichment in high-attention tiles, high-attention tiles were found in similar regions of the embedding space as tumor tiles.

We visualized randomly selected tiles from each attention score group (Figure 8b). Tiles in high-attention score groups were typically homogenous tumor regions, while tiles in low-attention groups were more variable, often including normal stromal cells, or empty or blurry spaces. These results suggest that the model relies strongly on tumor tiles for *MET* status prediction, while tiles capturing the context such as stroma, epithelium, and immune are of lower importance.

# Discussion

Protein overexpression of MET is a targetable event in NSCLC with drugs currently under development or in clinical trials. While overexpression is evaluable by IHC, NGS detection of overexpression would allow panel-based simultaneous detection of multiple actionable mutations in NSCLC. In this study, we developed a machine learning classifier based on H&E WSIs to predict the presence of *MET* overexpression and demonstrated model robustness to changes that may be seen in clinical data.

NGS of NSCLC tumors is increasingly performed as part of routine clinical care [21,22]. DNA-based NGS is common and RNA-based NGS is increasing in use as it offers the opportunity to identify actionable gene fusion events that would not otherwise have been found during DNA sequencing [10]. Recent studies have described an association of *MET* RNA expression with MET IHC staining [9,23]. If high concordance between RNA expression and IHC staining is demonstrated, RNA sequencing offers a single-assay companion diagnostic that detects *MET* overexpression simultaneously with actionable fusion events. In addition, measurement of RNA expression is not susceptible to discordances due to variance in pathologist IHC interpretation and could be used in cases where IHC staining is equivocal [24]. Future work should explore the RNA expression threshold for optimal *MET* overexpression therapy as this is not currently known.

Despite its advantages, NGS has limitations such as availability of tissue for sequencing and access to validated lab testing. Because generating H&E-stained slides is a common step in the diagnostic clinical workflow, a machine learning model trained to predict *MET* status directly from an H&E slide image could be used to prioritize patient tissue samples for further diagnostic workup. Our model demonstrated the ability to predict RNA-based *MET* status without using pixel-level annotations to learn a continuous correlation between predicted probability and *MET* RNA expression (Figure 5c and 6c). Once a therapy-defined RNA threshold is available, our model could be leveraged to stratify response to therapy meaningfully by thresholding model output probabilities accordingly. Model performance was observed to be similar across clinical subgroups, which is an important consideration for deployment. Finally, an evaluation of model performance prospectively over a 10-month period confirmed that our model's discrimination ability persists over time.

In a real-world setting, an imaging-based algorithm could encounter WSIs scanned with different devices resulting in visual differences in generated images. We evaluated model performance on WSIs obtained from a scanner different from that used for model development and observed only minimal performance degradation. We also observed consistent model performance in real-world stain variation in slides stained externally from our institution. We simulated the effect of stain differences on color by introducing synthetic hue, saturation, contrast, and brightness noise. We observed that performance as measured by rank-based metrics was stable, suggesting that the model maintains its discriminative ability in the face of such color alterations. Operating point-sensitive metrics such as sensitivity and specificity were generally stable at synthetic color alterations levels that reflect real-world-encountered visual differences, but broke down at higher synthetic levels. Future work to deploy this model for clinical use could address any concerns for operating point sensitivity using deployment site-specific determination of operating point.

Neural network approaches to prediction of labels from WSIs are thought of as "black-boxes", in which it is challenging to interpret how the deep learning model arrived at its decision. This can be a challenge for physician adoption of such models, as trust that the model has learned meaningful biological signals must be earned. We leveraged attention scatter plots generated by the model and region-level annotation of histological features as a means to get some insight into model behavior. As expected, we observed that tumor regions had higher attention scores relative to stromal, necrotic, epithelial, immune, and artifact areas.

The ultimate goal of the algorithm described here and other histogenomic algorithms is to incorporate them into the clinical workflow. Integration of AI models in the workflow could help clinicians by prioritizing patients with a high likelihood of a mutation for confirmatory testing. Following further optimization of model robustness, future work will focus on validating the model in real-world retrospective cohorts and prospective and interventional studies to evaluate clinical value.

# Data availability statement

Deidentified data used in the research was collected in a real-world health care setting and is subject to controlled access for privacy and proprietary reasons. When possible, derived data supporting the findings of this study have been made available within the paper and its Supplementary Figures/Tables.

# References

1. Sung H, Ferlay J, Siegel RL, et al. Global Cancer Statistics 2020: GLOBOCAN Estimates of Incidence and Mortality Worldwide for 36 Cancers in 185 Countries. *CA Cancer J Clin*. 2021;71(3):209-249. doi:10.3322/caac.21660

2. Yang SR, Schultheis AM, Yu H, Mandelker D, Ladanyi M, Büttner R. Precision medicine in non-small cell lung cancer: Current applications and future directions. *Semin Cancer Biol*. 2022;84:184-198. doi:10.1016/j.semcancer.2020.07.009

3. Landi L, Minuti G, D'Incecco A, Salvini J, Cappuzzo F. MET overexpression and gene amplification in NSCLC: a clinical perspective. *Lung Cancer Auckl NZ*. 2013;4:15-25. doi:10.2147/LCTT.S35168

4. Go H, Jeon YK, Park HJ, Sung SW, Seo JW, Chung DH. High MET Gene Copy Number Leads to Shorter Survival in Patients with Non-small Cell Lung Cancer. *J Thorac Oncol*. 2010;5(3):305-313. doi:10.1097/JTO.0b013e3181ce3d1d

5. Wolf J, Seto T, Han JY, et al. Capmatinib in MET Exon 14-Mutated or MET-Amplified Non-Small-Cell Lung Cancer. *N Engl J Med*. 2020;383(10):944-957. doi:10.1056/NEJMoa2002787

6. Camidge DR, Bar J, Horinouchi H, et al. Telisotuzumab vedotin (Teliso-V) monotherapy in patients (pts) with previously treated c-Met–overexpressing (OE) advanced non-small cell lung cancer (NSCLC). *J Clin Oncol*. 2022;40(16_suppl):9016-9016. doi:10.1200/JCO.2022.40.16_suppl.9016

7. Guo R, Berry LD, Aisner DL, et al. MET IHC is a Poor Screen for MET Amplification or MET exon 14 mutations in Lung Adenocarcinomas: Data from a Tri-Institutional Cohort of the Lung Cancer Mutation Consortium. *J Thorac Oncol Off Publ Int Assoc Study Lung Cancer*. 2019;14(9):1666-1671. doi:10.1016/j.jtho.2019.06.009

8. Tong JH, Yeung SF, Chan AWH, et al. MET Amplification and Exon 14 Splice Site Mutation Define Unique Molecular Subgroups of Non–Small Cell Lung Carcinoma with Poor Prognosis. *Clin Cancer Res*. 2016;22(12):3048-3056. doi:10.1158/1078-0432.CCR-15-2061

9. Bersani F, Picca F, Morena D, et al. Exploring circular MET RNA as a potential biomarker in tumors exhibiting high MET activity. *J Exp Clin Cancer Res CR*. 2023;42(1):120. doi:10.1186/s13046-023-02690-5

10. Michuda J, Park BH, Cummings AL, et al. Use of clinical RNA-sequencing in the detection of actionable fusions compared to DNA-sequencing alone. *J Clin Oncol*. 2022;40(16_suppl):3077-3077. doi:10.1200/JCO.2022.40.16_suppl.3077

11. Campanella G, Hanna MG, Geneslaw L, et al. Clinical-grade computational pathology using weakly supervised deep learning on whole slide images. *Nat Med*. 2019;25(8):1301-1309. doi:10.1038/s41591-019-0508-1

12. Lu MY, Williamson DFK, Chen TY, Chen RJ, Barbieri M, Mahmood F. Data-efficient and weakly supervised computational pathology on whole-slide images. *Nat Biomed Eng*. 2021;5(6):555-570. doi:10.1038/s41551-020-00682-w

13. Gagné A, Wang E, Bastien N, et al. Impact of Specimen Characteristics on PD-L1 Testing in Non–Small Cell Lung Cancer: Validation of the IASLC PD-L1 Testing Recommendations. *J Thorac Oncol*. 2019;14(12):2062-2070. doi:10.1016/j.jtho.2019.08.2503

14. Ronneberger O, Fischer P, Brox T. U-Net: Convolutional Networks for Biomedical Image Segmentation. In: Navab N, Hornegger J, Wells WM, Frangi AF, eds. *Medical Image Computing and Computer-Assisted Intervention – MICCAI 2015*. Lecture Notes in Computer

Science. Springer International Publishing; 2015:234-241. doi:10.1007/978-3-319-24574-4_28

15. Ilse M, Tomczak JM, Welling M. Attention-based Deep Multiple Instance Learning. Published online June 28, 2018. doi:10.48550/arXiv.1802.04712

16. He K, Zhang X, Ren S, Sun J. Deep Residual Learning for Image Recognition. Published online December 10, 2015. Accessed October 5, 2022. http://arxiv.org/abs/1512.03385

17. Kingma DP, Ba J. Adam: A Method for Stochastic Optimization. Published online January 29, 2017. Accessed October 5, 2022. http://arxiv.org/abs/1412.6980

18. Ingale K, Joshi RP, Ho IY, BenTaieb A, Stumpe MC. Effects of Color Calibration via ICC Profile on Inter-scanner Generalization of AI Models IN USCAP 2022 Abstracts: Informatics (977-1017). *Mod Pathol*. 2022;35(2):1163-1210. doi:10.1038/s41379-022-01042-6

19. Sha L, Osinski BL, Ho IY, et al. Multi-Field-of-View Deep Learning Model Predicts Nonsmall Cell Lung Cancer Programmed Death-Ligand 1 Status from Whole-Slide Hematoxylin and Eosin Images. *J Pathol Inform*. 2019;10(1):24. doi:10.4103/jpi.jpi_24_19

20. McInnes L, Healy J, Melville J. UMAP: Uniform Manifold Approximation and Projection for Dimension Reduction. Published online September 17, 2020. doi:10.48550/arXiv.1802.03426

21. Cainap C, Balacescu O, Cainap SS, Pop LA. Next Generation Sequencing Technology in Lung Cancer Diagnosis. *Biology*. 2021;10(9):864. doi:10.3390/biology10090864

22. Smeltzer MP, Wynes MW, Lantuejoul S, et al. The International Association for the Study of Lung Cancer Global Survey on Molecular Testing in Lung Cancer. *J Thorac Oncol*. 2020;15(9):1434-1448. doi:10.1016/j.jtho.2020.05.002

23. Aguado C, Teixido C, Román R, et al. Multiplex RNA-based detection of clinically relevant MET alterations in advanced non-small cell lung cancer. *Mol Oncol*. 2021;15(2):350-363. doi:10.1002/1878-0261.12861

24. Imyanitov EN, Ivantsov AO, Tsimafeyeu IV. Harmonization of Molecular Testing for Non-Small Cell Lung Cancer: Emphasis on PD-L1. *Front Oncol*. 2020;10:549198. doi:10.3389/fonc.2020.549198

# Figures Legends

**Figure 1. Schematic representation of study.**

(a) The study was set up to compare RNA expression against DNA amplification and wildtype cases to determine threshold for RNA expression status. (b) An attention-based multiple instance learning model was trained to predict MET status from H&E images.

**Figure 2. Machine learning experiments results**.

(a) ROC curve for test folds in cross validation. (b) Scatterplot compares predicted probabilities from the model against RNA expression values. (c) Boxplots comparing models trained with tiles at 5x, 10x and 20x magnification. Models trained with 20x had a significantly better ROC-AUC on test folds than other magnifications. (d) Boxplots comparing model performance at different dataset sizes. Marginal model performance improvement might be observed with more *MET*-positive cases in the training dataset.

**Figure 3. Results on independent holdout test set scanned with Philips UFS scanner.**

(a) ROC curve for holdout test set. (b) Precision-recall curve for holdout test set. (c) Correlation plot between RNA expression values and model-predicted probability. (d) Operating point based metrics on holdout test set.

**Figure 4. Receiver operating characteristic (ROC) curves for different clinical subgroups.**

(a) ROC curves stratified by clinical subgroups. Significant differences were found between large vs small specimens and current smokers vs ex-smokers. Other subgroups did not show statistically significant differences. (b) ROC curves with other mutation status as labels and model-predicted probabilities as scores. ROC-AUC values close to 0.5 indicate the model has not learned features for these other mutations.

**Figure 5. Adding synthetic noise to test set**

(a) Metrics as a function of color perturbation level. Rank-based metrics like ROC-AUC seem to be stable, but shifts in probability distributions result in diverging operating point-based metrics.

(b) Examples of color perturbation on a test image with increasing level of perturbation resulting in perturbation factor sampled from a larger interval.

**Figure 6. Independent holdout test set scanned with an Aperio GT450 scanner**

(a) ROC curve. (b) Precision-recall curve. (c) Correlation plot between RNA expression values and model-predicted probability. (d) Operating point-based metrics on the holdout test set scanned with an Aperio GT450 scanner. Model performance is similar to the Philips UFS scanner test set in terms of ROC-AUC and average precision. However, differences in sensitivity and specificity compared to the Philips UFS scanner test set indicate a shift in predicted probability distribution.

**Figure 7. Model performance on the temporal test set.**

Receiver operating characteristic (ROC) curves for (a) cases stained externally and scanned with a Philips scanner, (b) cases stained internally and scanned with a Philips scanner. (c) cases stained externally and scanned with an Aperio scanner, and (d) cases stained internally and scanned with an Aperio scanner.

**Figure 8: Model interpretability.**

a) Histology class prevalence in high-attention (importance) tiles (tiles accounting for the top 10% of the cumulative attention probability). Tiles from each WSI were classified into 6 classes. The fraction of tiles of a histological class for all tiles and for high-attention tiles in each slide is shown (N=588). b) Distribution of tiles in the embedding space and tile mosaics. UMAPs of tile embedding vectors are at the center. Attention Score Groups represent the percentile of attention score of all tiles. Randomly selected tiles from attention score groups are shown on the left and right.

## Acknowledgements

We thank Matthew Kase for a rigorous review of the manuscript and Evan Gendell for figure design. We thank Andrew Bandy, Andrey Khramtsov, Arash Mohtashamian, Cristina Isales and Elias Makhoul for dataset annotations. We thank Andrew Kruger for contributions to code infrastructure.

## Author contributions

A.B., J.S.K.B. and R.P.J. performed study concept and design; L.S., I.H., K.I., S.H. and R.P.J. performed development of methodology; K.I., S.H., K.N., R.P.J. and M.C.S. contributed to writing, review and revision of the paper; K.I., S.H., A.R., K.N. and R.P.J. provided analysis and interpretation of data; A.W. provided technical and material support. All authors read and approved the final paper.

## Funding

This work was supported by Tempus Labs.

# Figures

**Figure 1. Schematic representation of study.**

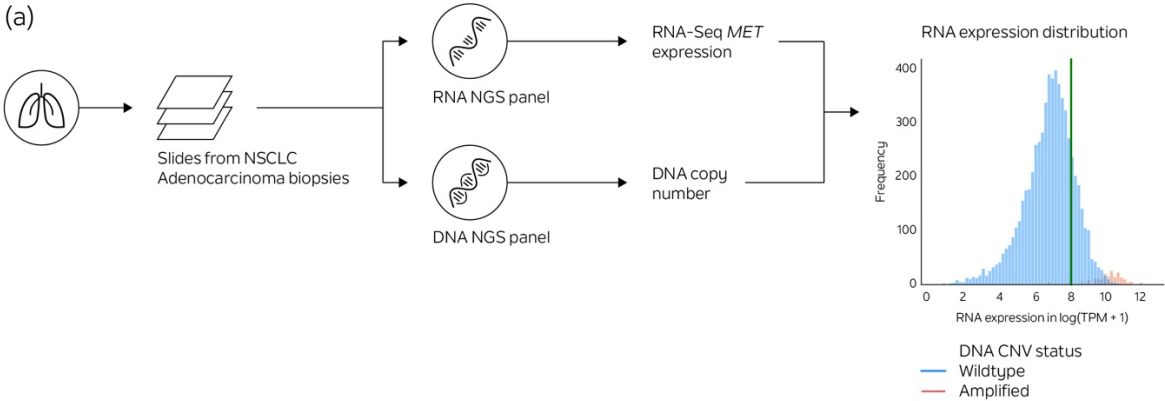

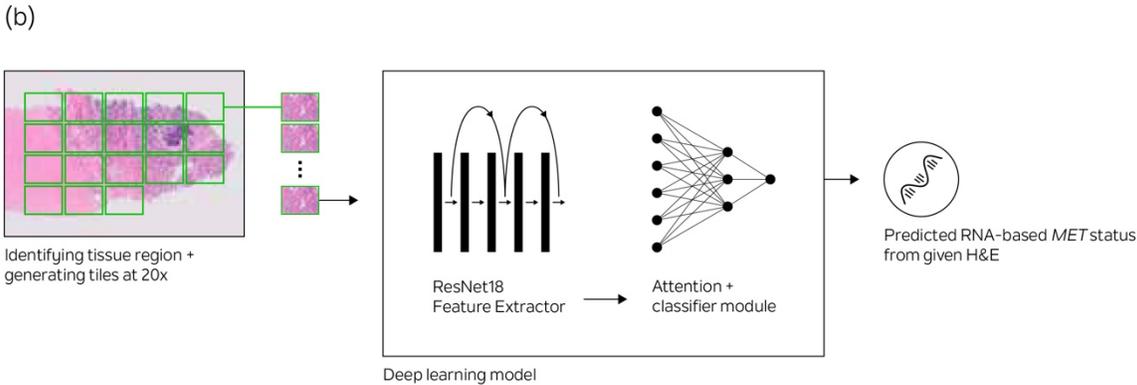

**Figure 2. Machine learning experiments results.**

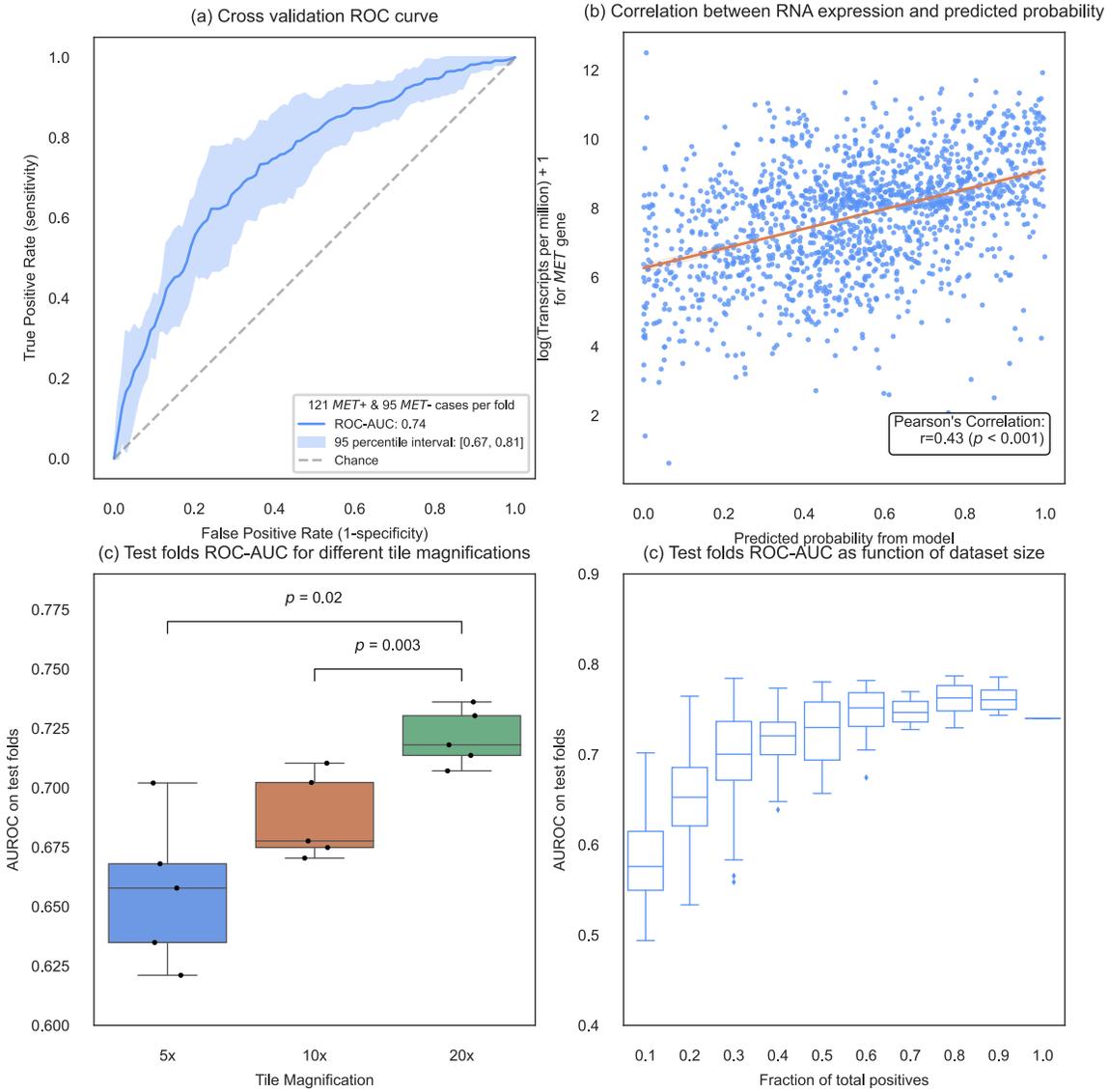

**Figure 3. Results on independent holdout test set scanned with Philips UFS scanner.**

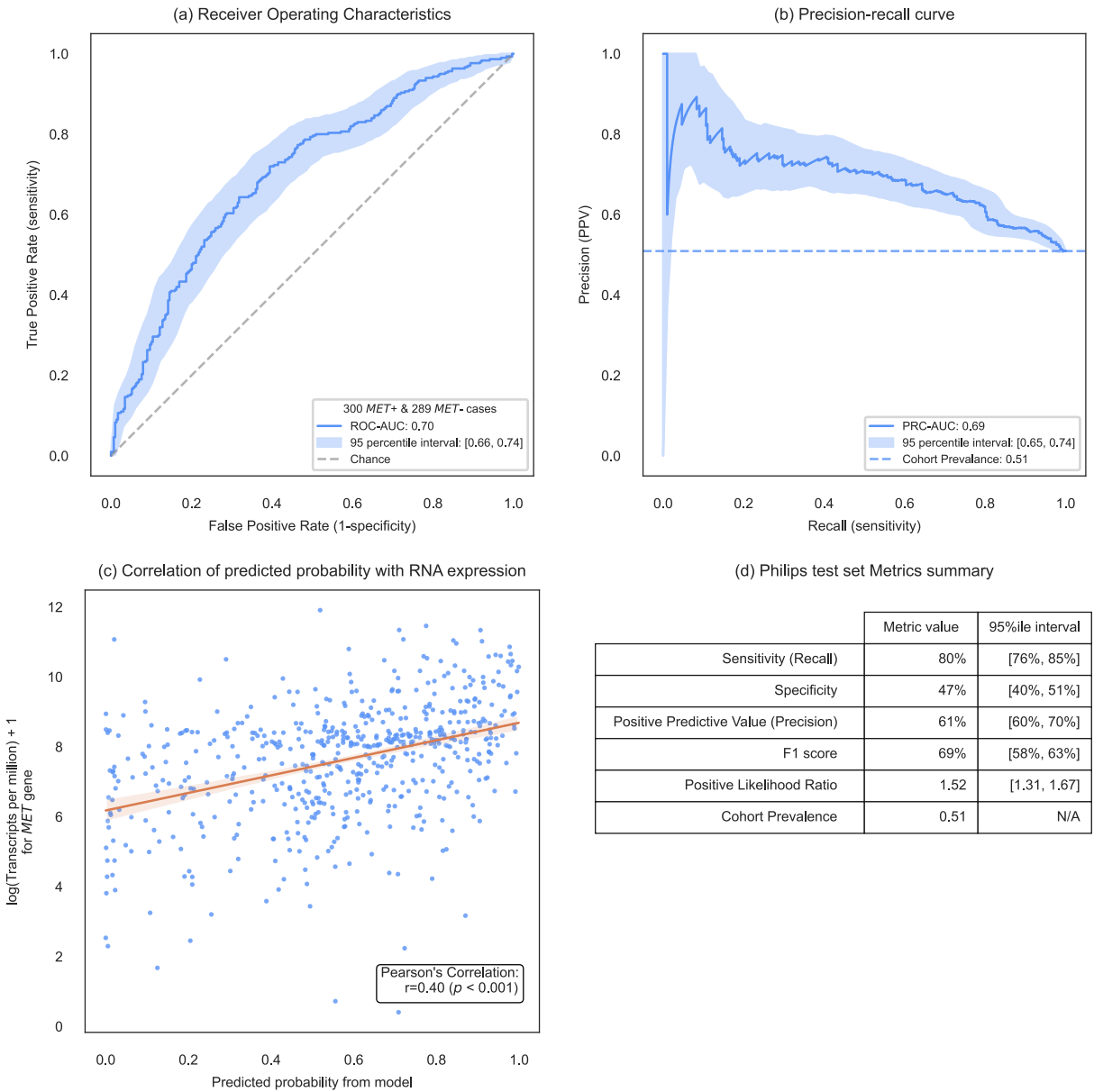

**Figure 4. Receiver operating characteristic (ROC) curves for different clinical subgroups.**

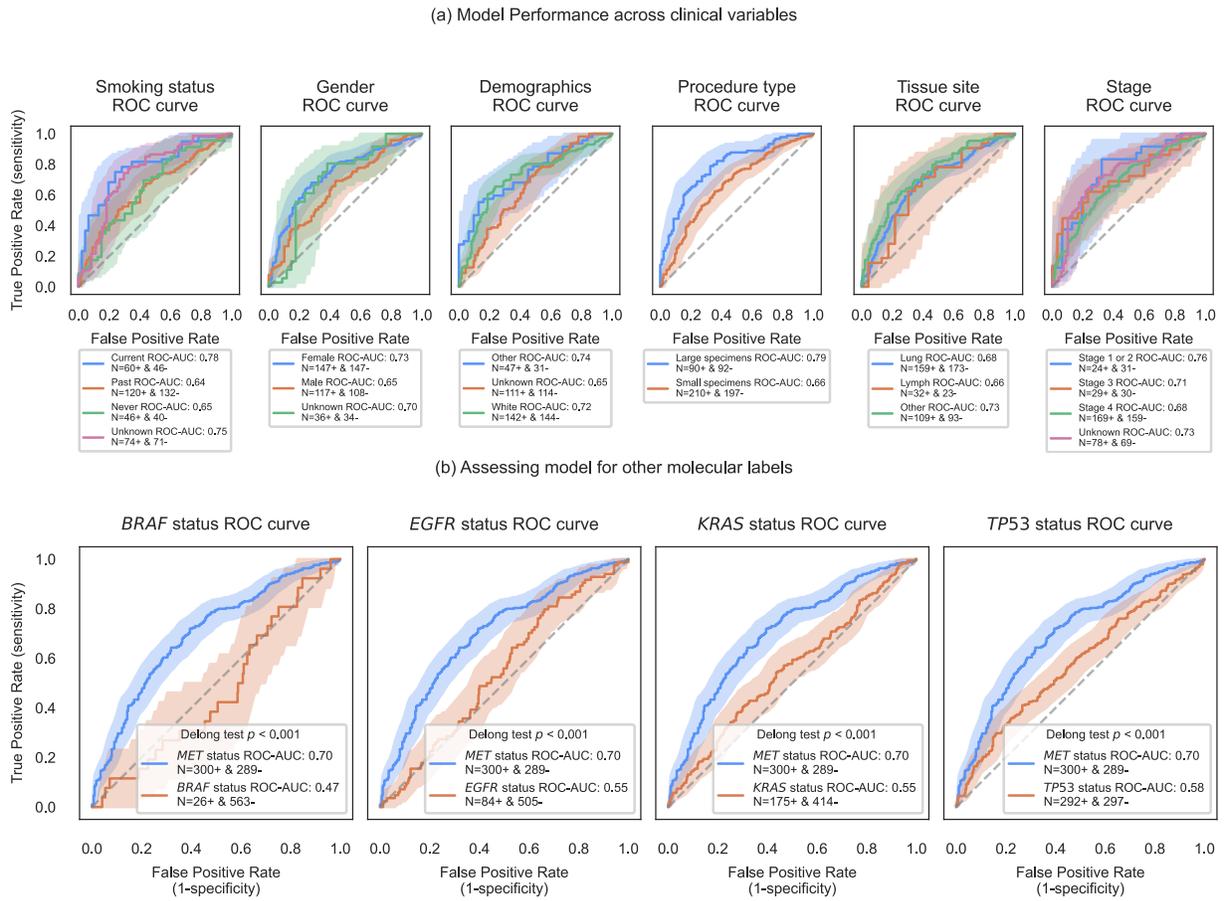

(a) Model Performance across clinical variables

(b) Assessing model for other molecular labels

**Figure 5. Adding synthetic noise to test set**

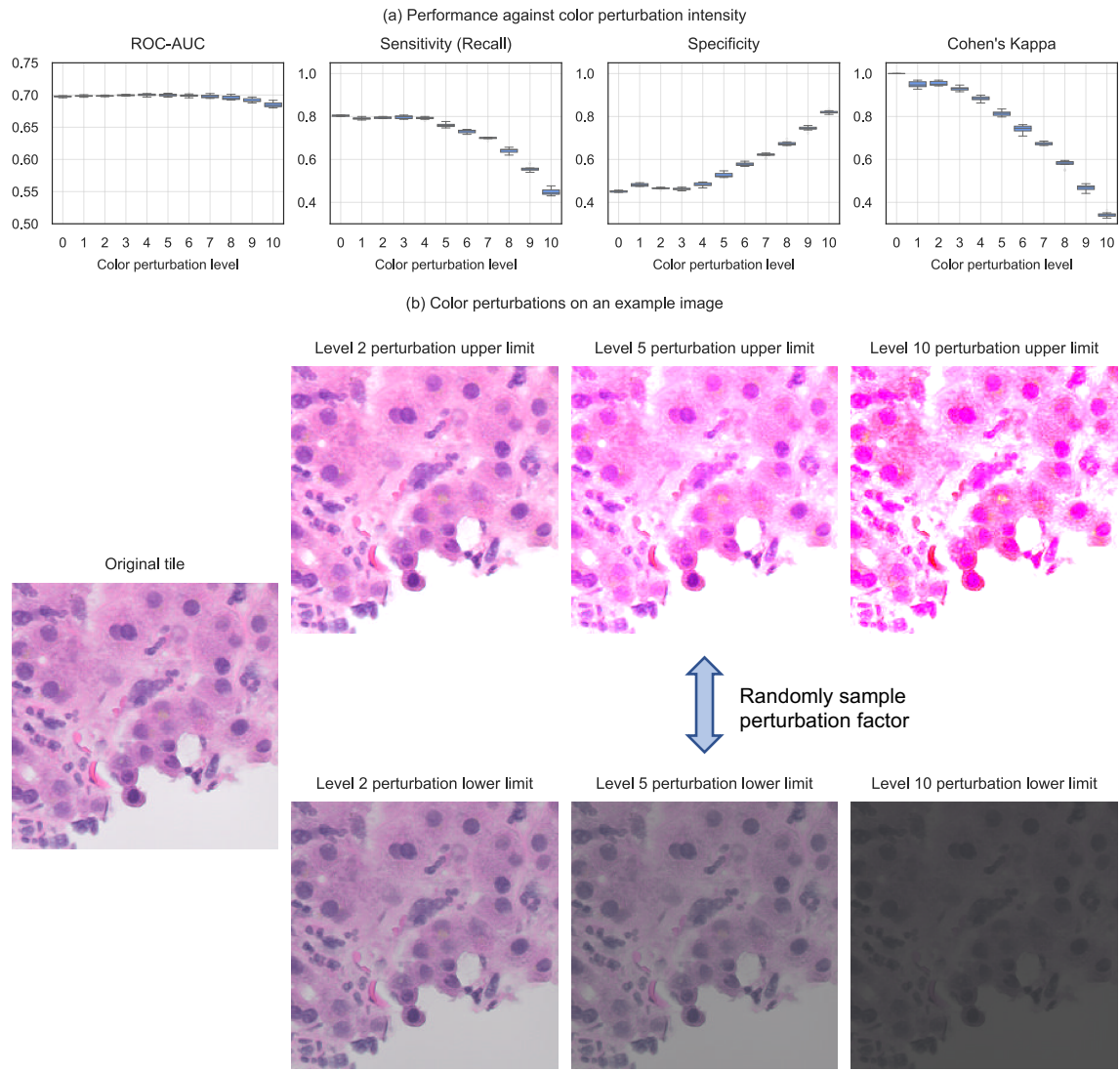

**Figure 6. Independent holdout test set scanned with an Aperio GT450 scanner**

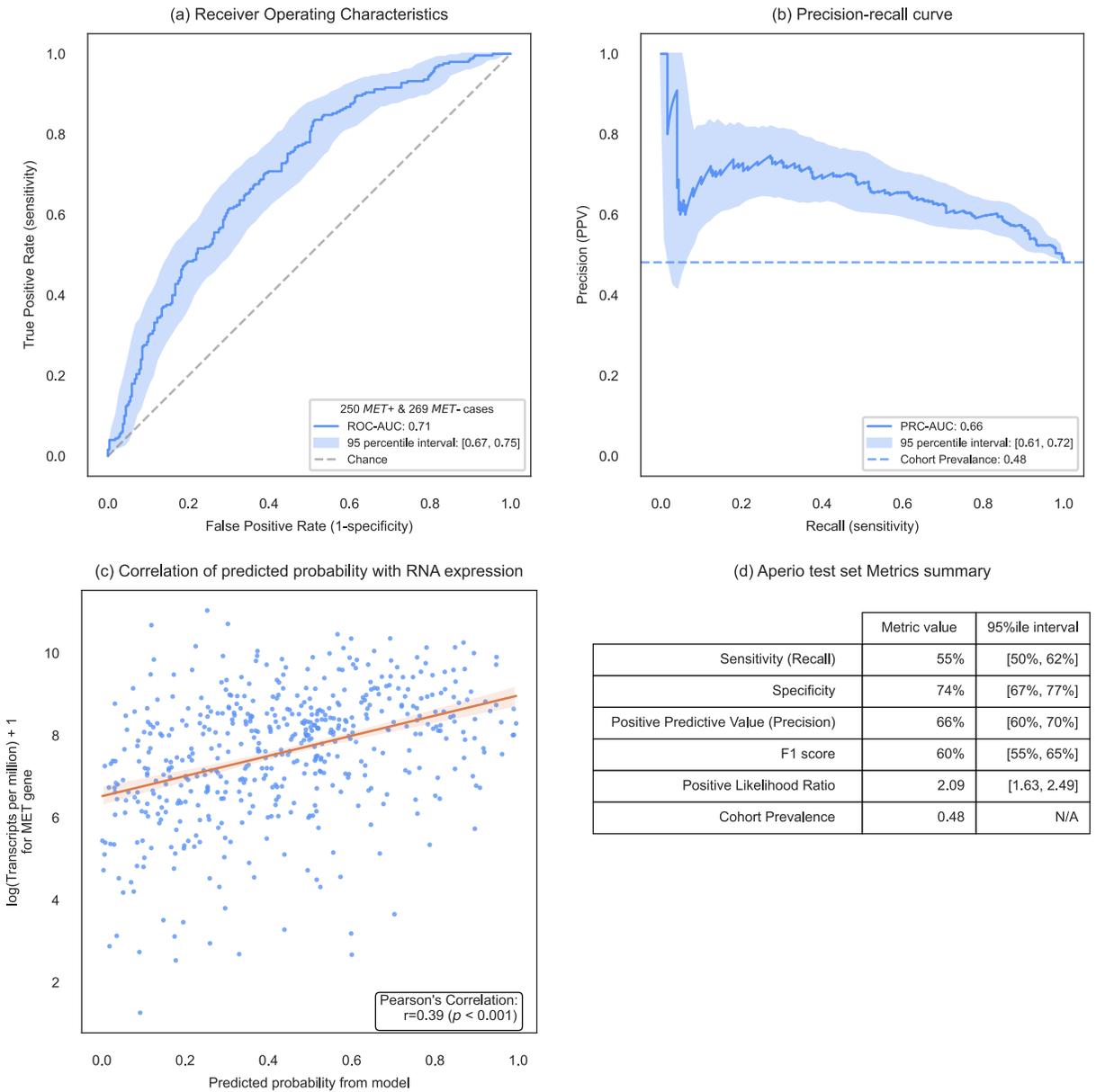

**Figure 7. Model performance on the temporal test set.**

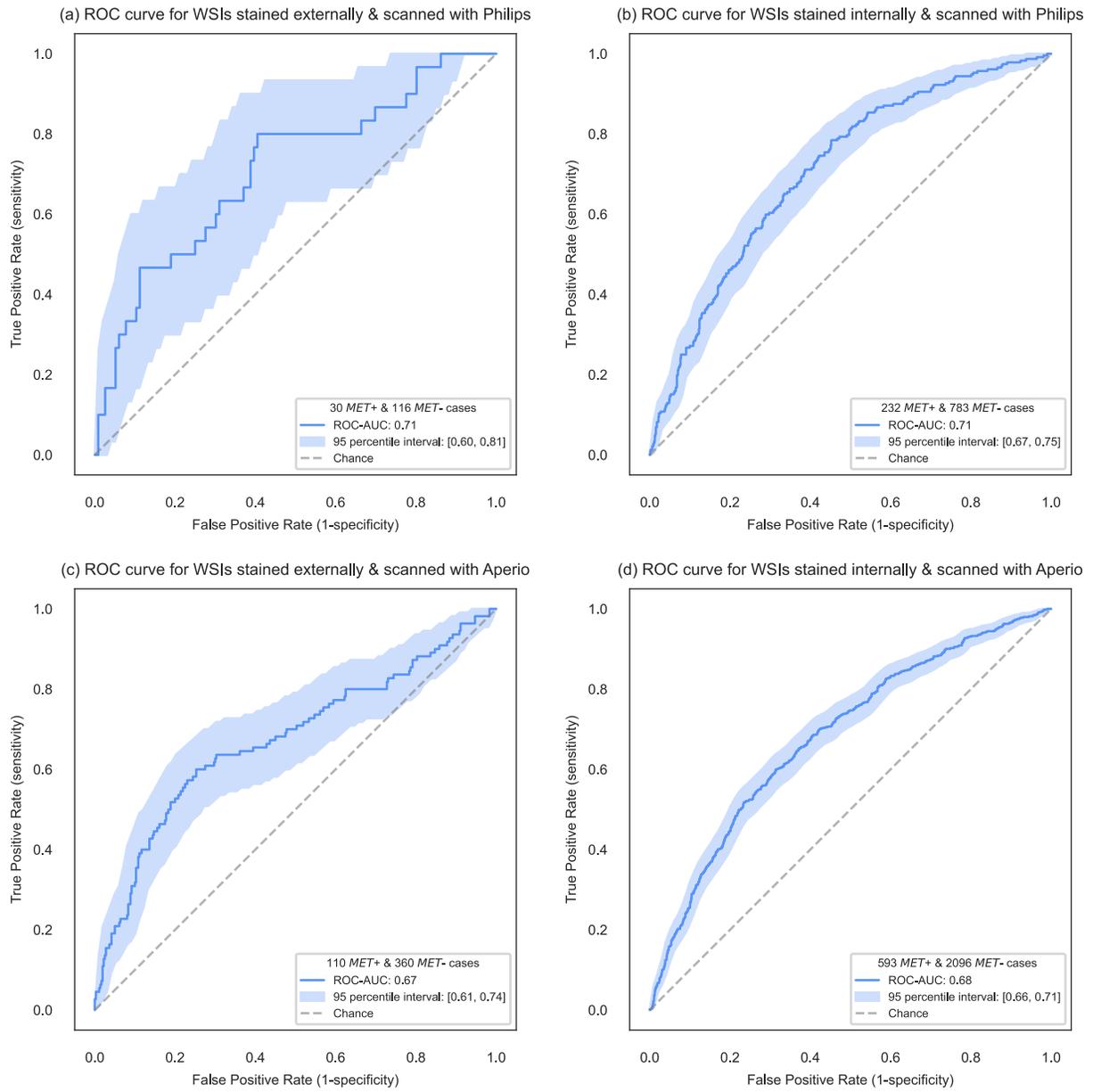

**Figure 8: Model interpretability.**

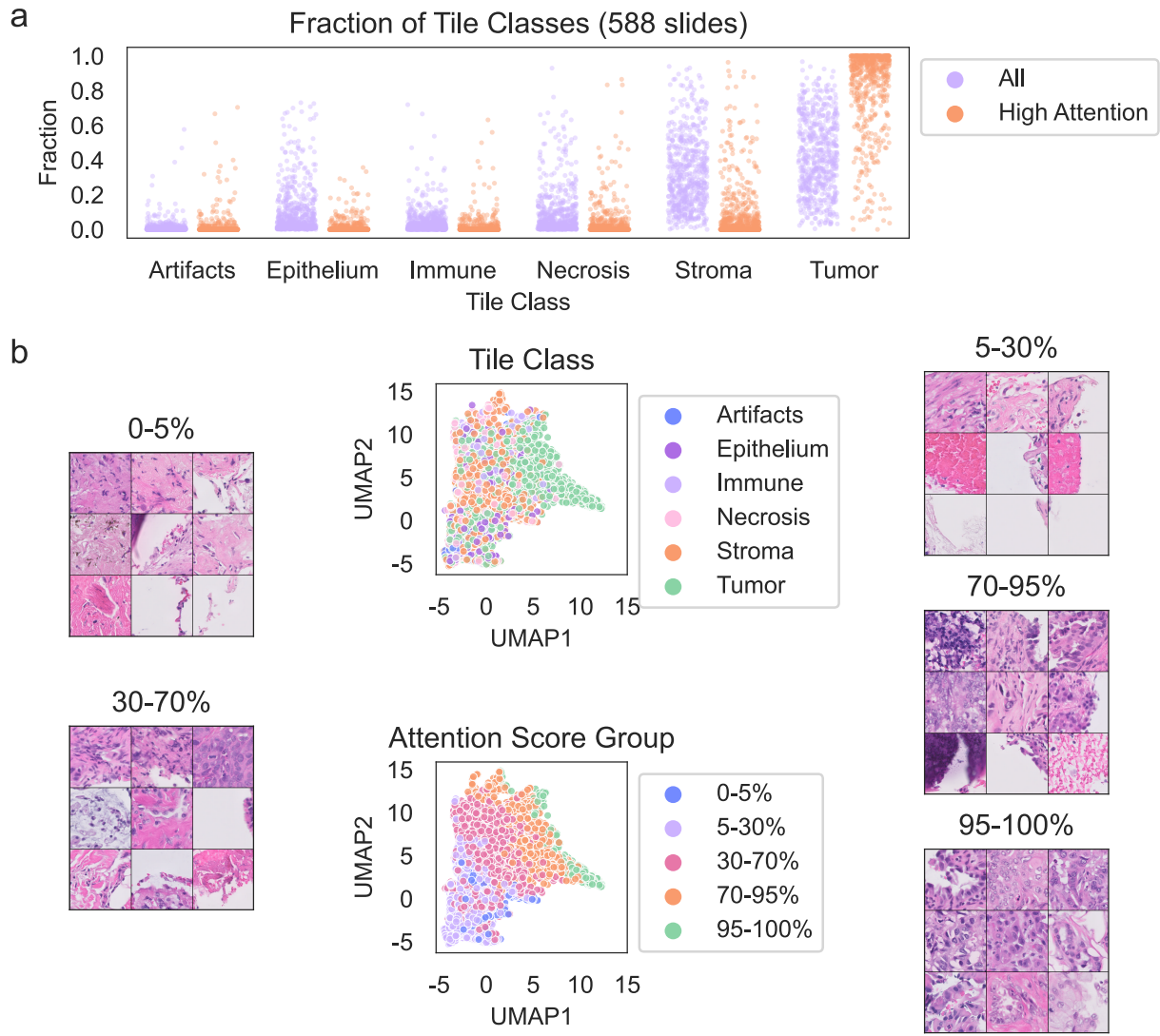

# Supplementary Images

**Supplementary Figure A1.** Histogram showing distribution of RNA expression values across the entire cohort and cohort sampled for model development. Model development cohort was enriched for sufficient positive cases.

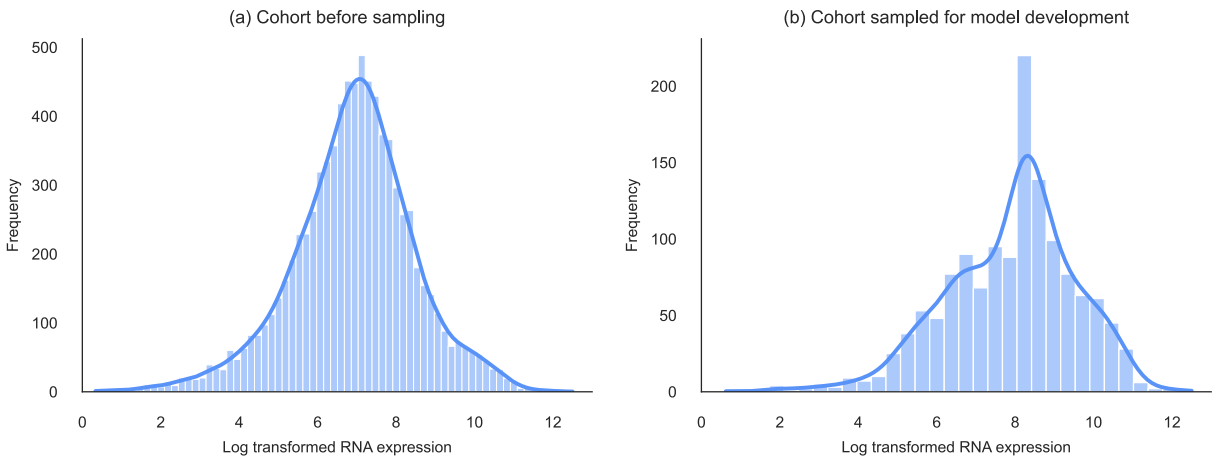

**Supplementary Figure A2.** Receiver operating characteristic curve with labels identified by DNA copy number >5 as positive and DNA copy number equal to 2 as negative and scores obtained by RNA expression values.

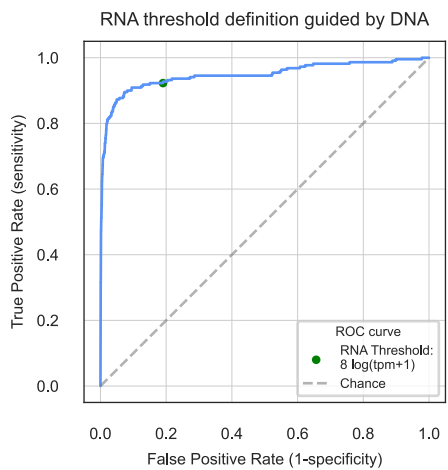

**Supplementary Figure A3.** Predicted probabilities from model trained with binary and smooth labels. Correlation between RNA overexpression and model predicted probability is slightly better with smooth labels.

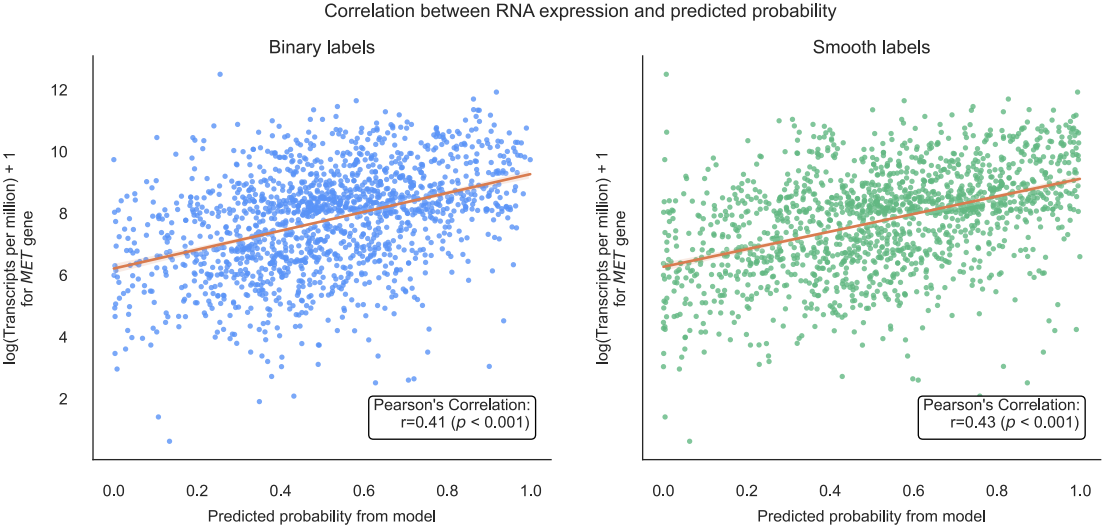

**Supplementary Figure A4.** Cross-validation performance on models trained and evaluated on slides digitized with both Aperio and Phillips scanners.

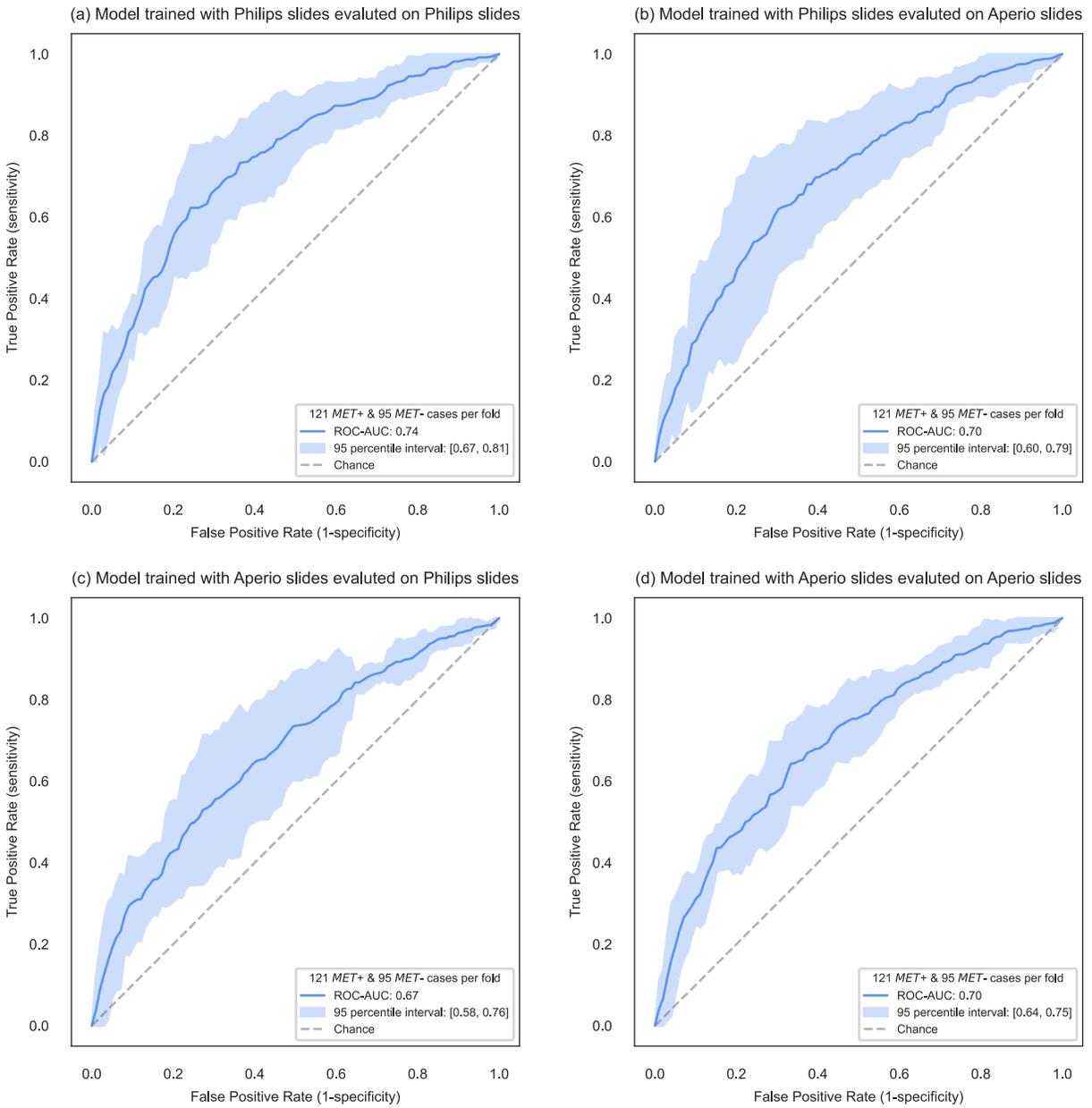

**Supplementary Figure A5.** Color robustness metrics and predicted probabilities (a) Rank-based metrics like average precision seem to be stable but shifts in probability distributions result in diverging operating point-based metrics. (b) On comparing test set (left) probabilities with the perturbed test set (right), synthetic noise in brightness, contrast, hue, and saturation drives the distribution to lower values.

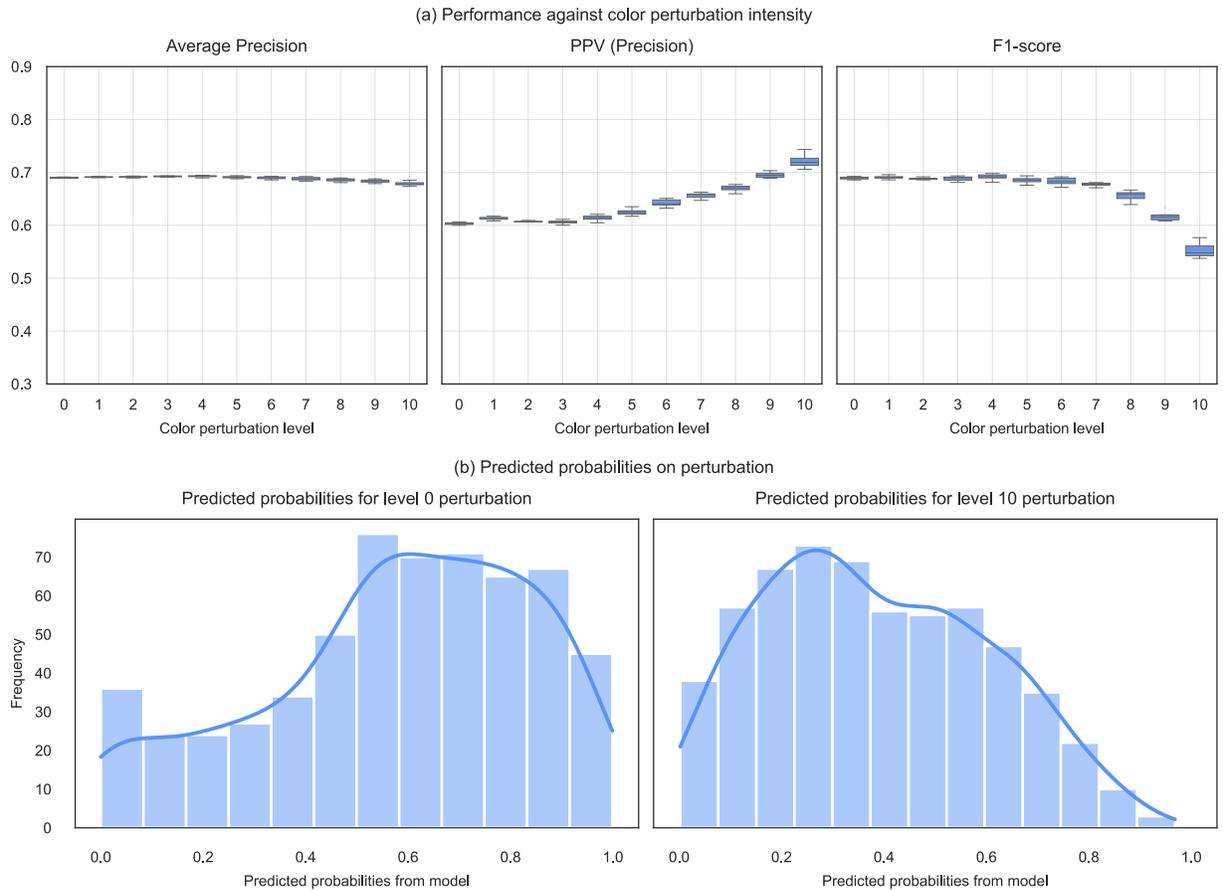

**Supplementary Figure A6.** Predicted probabilities from the model for the Philips test set and Aperio test set. Predicted probabilities appear to shift when inferred on images obtained from different scanners (P < 0.01).

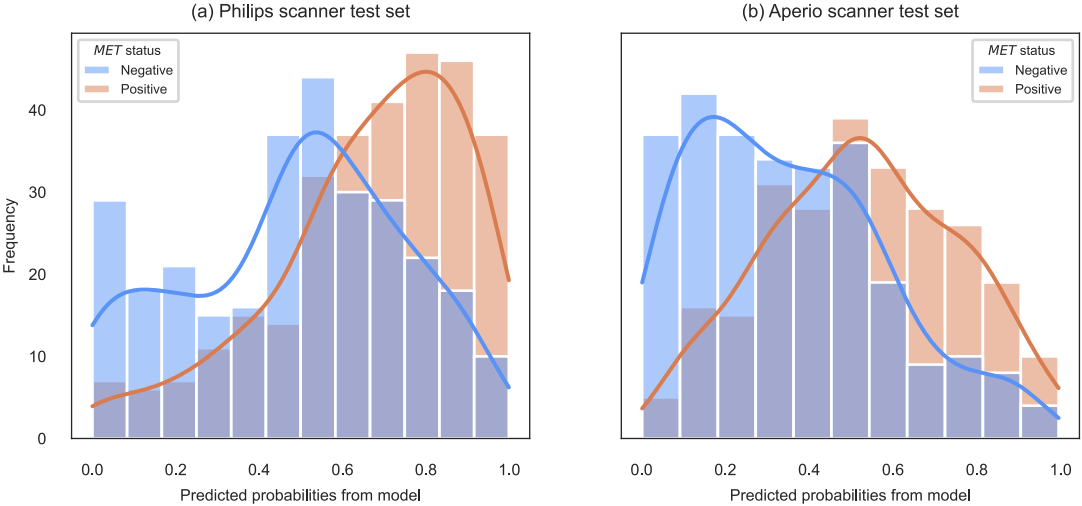

# Supplementary Tables

**Supplementary Table 1.** Model development and holdout test set cohort characteristics. PD-L1 status indicated a statistically significant association with MET labels.

| Characteristics | Model development set | | | Holdout test set | | |
|---|---|---|---|---|---|---|
| | *MET normal expression* | *MET overexpression* | *p* | *MET normal expression* | *MET overexpression* | *p* |
| **Has BRAF alteration** | | | | | | |
| Negative | 579 | 720 | 0.02 | 274 | 289 | 0.48 |
| Positive | 20 | 48 | | 15 | 11 | |
| **Has EGFR alteration** | | | | | | |
| Negative | 494 | 635 | 0.98 | 250 | 255 | 0.69 |
| Positive | 105 | 133 | | 39 | 45 | |
| **Has KRAS alteration** | | | | | | |
| Negative | 410 | 526 | 1 | 209 | 205 | 0.33 |
| Positive | 189 | 242 | | 80 | 95 | |
| **Has TP53 alteration** | | | | | | |
| Negative | 290 | 286 | < 0.001 | 149 | 148 | 0.65 |
| Positive | 309 | 482 | | 140 | 152 | |
| **Smoking status** | | | | | | |
| Current smoker | 90 | 138 | 0.11 | 46 | 60 | 0.44 |
| Ex-smoker | 272 | 303 | | 132 | 120 | |
| Never smoker | 83 | 104 | | 40 | 46 | |
| Unknown | 154 | 223 | | 71 | 74 | |
| **Gender** | | | | | | |
| Female | 283 | 347 | 0.45 | 147 | 147 | 0.9 |
| Male | 227 | 288 | | 108 | 117 | |
| Unknown | 89 | 133 | | 34 | 36 | |
| **Race** | | | | | | |
| Asian | 15 | 12 | 0.68 | 8 | 14 | 0.31 |
| Black or African American | 44 | 63 | | 17 | 19 | |
| Other Race | 12 | 16 | | 6 | 14 | |
| Unknown | 254 | 339 | | 114 | 111 | |
| White | 274 | 338 | | 144 | 142 | |
| **Stage** | | | | | | |
| Stage 0 | 0 | 1 | 0.08 | 0 | 0 | 0.42 |
| Stage 1 | 40 | 61 | | 17 | 18 | |
| Stage 2 | 43 | 28 | | 14 | 6 | |
| Stage 3 | 57 | 71 | | 30 | 29 | |
| Stage 4 | 296 | 386 | | 159 | 169 | |
| Unknown | 163 | 221 | | 69 | 78 | |
| **Tissue site** | | | | | | |
| lung | 362 | 394 | 0.006 | 173 | 159 | 0.21 |
| lymph | 50 | 91 | | 23 | 32 | |
| missing | 5 | 5 | | 0 | 0 | |
| other | 182 | 278 | | 93 | 109 | |
| **Specimen size** | | | | | | |
| Large specimens | 203 | 219 | 0.1 | 92 | 90 | 0.69 |
| **Small specimens** | 385 | 535 | | 197 | 210 | |
| **Unknown** | 11 | 14 | | 0 | 0 | |
| **PDL1 status** | | | | | | |
| **Negative** | 199 | 118 | < 0.001 | 117 | 52 | < 0.001 |
| **Positive** | 222 | 444 | | 103 | 178 | |
| **Unknown** | 178 | 206 | | 69 | 70 | |

**Supplementary Table 2.** The fraction of tiles of a histological class for all tiles and for high attention tiles in each slide (n=588). Wilcoxon signed rank sum tests were performed to test if a tile class has higher or lower prevalence among high attention tiles. The p-values are corrected with the Benjamini/Hochberg False Discovery Rate (FDR) correction method.

| Tile Class | All Tile Fraction | High Attention Tile Fraction | Alternative Hypothesis | FDR Corrected Wilcoxon p-value |
|---|---|---|---|---|
| Tumor | 0.423 ± 0.217 | 0.828 ± 0.230 | Greater In High Attention Tiles | < 0.001 |
| Stroma | 0.321 ± 0.204 | 0.092 ± 0.169 | Less In High Attention Tiles | < 0.001 |
| Necrosis | 0.095 ± 0.138 | 0.036 ± 0.099 | Less In High Attention Tiles | < 0.001 |
| Immune | 0.044 ± 0.073 | 0.016 ± 0.057 | Less In High Attention Tiles | < 0.001 |
| Artifacts | 0.013 ± 0.042 | 0.015 ± 0.063 | Greater In High Attention Tiles | 1 |
| Epithelium | 0.104 ± 0.141 | 0.014 ± 0.040 | Less In High Attention Tiles | < 0.001 |

**Supplementary Table 3.** Fraction of tile classes among high attention tiles were calculated in each slide. Tile fraction column shows mean and standard deviation over all slides (n=588). Wilcoxon signed rank sum tests were performed to test if tumor tiles are more prevalent than other tile classes. The p-values are corrected with the Benjamini/Hochberg False Discovery Rate (FDR) correction method.

| Tile Class | Tile Fraction (mean ± SD) | FDR Corrected p-value |
|---|---|---|
| **Tumor** | 0.828 ± 0.230 | NA |
| **Stroma** | 0.092 ± 0.169 | < 0.001 |
| **Necrosis** | 0.036 ± 0.099 | < 0.001 |
| **Immune** | 0.016 ± 0.057 | < 0.001 |
| **Artifacts** | 0.015 ± 0.063 | < 0.001 |
| **Epithelium** | 0.014 ± 0.040 | < 0.001 |

**Supplementary table 4.** Cohort characteristics for generalization and temporal test set

| Characteristics | Aperio scanner generalization set | | | Temporal generalization set | | |
|---|---|---|---|---|---|---|
| | *MET normal expression* | *MET overexpression* | *p value* | *MET normal expression* | *MET overexpression* | *p value* |
| **Has BRAF alteration** | | | | | | |
| **Negative** | 262 | 233 | 0.04 | 3219 | 907 | 0.004 |
| **Positive** | 7 | 17 | | 142 | 63 | |
| **Has EGFR alteration** | | | | | | |
| **Negative** | 224 | 214 | 0.54 | 2830 | 806 | 0.44 |
| **Positive** | 45 | 36 | | 531 | 164 | |
| **Has KRAS alteration** | | | | | | |
| **Negative** | 182 | 153 | 0.15 | 2221 | 573 | < 0.001 |
| **Positive** | 87 | 97 | | 1140 | 397 | |
| **Has TP53 alteration** | | | | | | |
| **Negative** | 109 | 90 | 0.33 | 1373 | 306 | < 0.001 |
| **Positive** | 160 | 160 | | 1988 | 664 | |
| **Smoking status** | | | | | | |
| **Current smoker** | 35 | 44 | 0.43 | 474 | 132 | 0.59 |
| **Ex-smoker** | 98 | 94 | | 899 | 241 | |
| **Never smoker** | 27 | 21 | | 307 | 93 | |
| **Unknown** | 109 | 91 | | 1681 | 504 | |
| **Gender** | | | | | | |
| **Female** | 106 | 103 | 0.83 | 1715 | 494 | 0.57 |
| **Male** | 88 | 83 | | 1631 | 469 | |
| **Unknown** | 75 | 64 | | 15 | 7 | |
| **Race** | | | | | | |
| Asian | 4 | 5 | 0.73 | 51 | 18 | 0.81 |
| Black or African American | 18 | 12 | | 139 | 35 | |
| Other Race | 3 | 1 | | 64 | 15 | |
| Unknown | 155 | 144 | | 2112 | 614 | |
| White | 89 | 88 | | 995 | 288 | |
| **Stage** | | | | | | |
| Stage 0 | 0 | 0 | 0.21 | 1 | 0 | 0.46 |
| Stage 1 | 16 | 12 | | 165 | 38 | |
| Stage 2 | 7 | 17 | | 111 | 24 | |
| Stage 3 | 26 | 22 | | 237 | 71 | |
| Stage 4 | 112 | 108 | | 1073 | 330 | |
| Unknown | 108 | 91 | | 1774 | 507 | |
| **Tissue site** | | | | | | |
| lung | 153 | 124 | 0.01 | 1990 | 526 | < 0.001 |
| lymph | 24 | 32 | | 306 | 131 | |
| missing | 7 | 0 | | 63 | 14 | |
| other | 85 | 94 | | 1002 | 299 | |
| **Specimen size** | | | | | | |
| Large specimens | 77 | 66 | 0.13 | 589 | 128 | 0.006 |
| **Small specimens** | 174 | 176 | | 2759 | 839 | |
| **Unknown** | 18 | 8 | | 13 | 3 | |
| **PDL1 status** | | | | | | |
| **Negative** | 89 | 31 | < 0.001 | 1333 | 164 | < 0.001 |
| **Positive** | 107 | 158 | | 1264 | 590 | |
| **Unknown** | 73 | 61 | | 764 | 216 | |